\documentclass{article}

\usepackage{graphicx} 
\usepackage{subfigure} 
\usepackage{microtype}
\graphicspath{{./}{./figures/}}

\usepackage{algorithm}
\usepackage[noend]{algorithmic}

\usepackage{amsmath}
\usepackage{amsfonts}
\usepackage{booktabs}
\usepackage{tabularx}
\usepackage{paralist}
\usepackage[T1]{fontenc}
\usepackage[utf8]{inputenc}
\usepackage{authblk}

\newcommand{\alg}[1]{\textsc{#1}}

\usepackage{dgleich-math}
\usepackage{amssymb}

\newcommand{\norm}[1]{\left\lVert#1\right\rVert}
\begin{document}

\title{Correlation Clustering with Low-Rank Matrices}

\author[1]{Nate Veldt}
\author[2]{Anthony Wirth}
\author[3]{David Gleich}
\affil[1]{Department of Mathematics, Purdue University}
\affil[2]{Department of Computer Science, Purdue University}
\affil[3]{Department of Computing and Information Systems, The University of Melbourne}
\renewcommand\Authands{ and }

\newcommand{\tony}[1]{{\color{Mulberry}{\bf{Tony says:}} \emph{#1}}}
\newcommand{\recentedit}[1]{{\color{Blue} {#1}}}
\maketitle
\begin{abstract}
	Correlation clustering is a technique for aggregating data based on
	qualitative information about which pairs of objects are labeled `similar' or `dissimilar.'
	Because the optimization problem is NP-hard, much of the previous literature
	focuses on finding approximation algorithms. In this paper we explore how to solve the correlation clustering objective exactly when the data to be clustered can be represented by a low-rank matrix. We prove in particular that correlation clustering can be solved in polynomial time when the underlying matrix is positive semidefinite
	with small constant rank, but that the task remains NP-hard in the presence of even one negative eigenvalue. Based on our theoretical results, we develop an
	algorithm for efficiently ``solving'' low-rank positive semidefinite correlation
	clustering by employing a procedure for zonotope vertex enumeration.
	We demonstrate the effectiveness and speed of our algorithm by using it to solve several clustering problems on both synthetic and real-world data. 
\end{abstract}

\section{Introduction}
Correlation clustering is a method for partitioning a dataset based on
pairwise information that indicates whether pairs of objects in the given dataset are `similar' or `dissimilar.' 
Typically correlation clustering is cast as a graph optimization problem
where the nodes of a graph represent objects from the dataset. In its most
basic form, the graph is assumed to be complete and unweighted, with each edge
being labeled `$+$' or `$-$' depending on whether the two nodes are `similar'
or `dissimilar.' Given this input, the objective is to partition the graph in
a way that maximizes the number of agreements, where an agreement is a `$+$'
edge that is included inside a cluster, or a `$-$' edge that links nodes in
different clusters. An equivalent objective, though more difficult to
approximate, is the goal of minimizing disagreements, i.e., `similar' nodes
that are separated or `dissimilar' nodes that are clustered together. A more
general form of correlation clustering associates each pair of objects
with not only a label but also a weight indicating how similar or dissimilar the
two objects are. In this case, the goal is to maximize the weight of agreements or minimize the weight of disagreements.

One attractive property of this clustering approach is that the number of
clusters formed is determined automatically by optimizing the objective function, rather than being a required input. In practice, correlation clustering has been applied in a wide variety of disciplines to solve problems such as cross-lingual link detection~\cite{van2007correlation}, gene clustering~\cite{bhattacharya2008divisive}, image segmentation~\cite{kim2011higher}, and record linkage in natural language
processing~\cite{NIPS2004_2557}.

Because correlation clustering is NP-hard~\cite{bansal2004correlation}, much of the previous literature has focused on developing approximation algorithms.
In this paper, we consider a new approach, exploring instances of the problem where the weighted labels can be represented by a low-rank matrix. Studying this case provides a new means for dealing with the intractability of the problem, and also allows us to apply the framework of correlation clustering in the broader task of analyzing low-dimensional datasets.

\textbf{Our Contributions}:
In this paper we prove that correlation clustering can be solved in polynomial
time when the similarity labels can be represented by a positive semidefinite
matrix of low rank. We also show that the problem remains NP-hard when the
underlying matrix has even one negative eigenvalue. To solve correlation
clustering problems in practice, we implement an algorithm called \alg{ZonoCC}
based on the randomized zonotope vertex enumeration procedure of Stinson,
Gleich, and Constantine~\cite{stinson2016randomized}.
This algorithm is capable of optimally solving
low-rank positive semidefinite correlation clustering. It is most useful, however, when it is truncated at a fixed number of iterations in order to quickly
obtain a very good approximation -- sans formal guarantee --  to the optimal solution. We demonstrate the effectiveness of \alg{ZonoCC} by obtaining clusterings for both synthetic and real-world datasets, including social network datasets and search-query data for well-known computer science conferences.
\section{Problem Statement}

We begin with the standard approach to correlation clustering by considering a
graph with $n$ nodes where edges are labeled either `$+$' or `$-$'.
Typically the correlation clustering objective is cast as an integer linear
program in the following way. For every pair of nodes $i$ and $j$ we are given
two nonnegative weights, $w_{ij}^+$ and $w_{ij}^-$, which indicate a score for
how similar the two nodes are and a score for how dissimilar they are
respectively. Traditionally, we assume that only one of these weights 
is nonzero (if not, they can be adjusted so this is the case without
changing the objective function by more than an additive constant). 
For every pair of nodes $i,j$ we introduce a binary variable $d_{ij}$ such that
\[ 
d_{ij} = \begin{cases}
0 & \text{ if $i$ and $j$ are clustered together}\,; \\
1 & \text{ if $i$ and $j$ are separated}\,.
\end{cases}
\]
In other words, $d_{ij} = 1$ indicates we have cut the edge between nodes $i$ and
$j$. The maximization version of correlation clustering is
given by the following ILP (integer linear program).
We include triangle constraints on the $d_{ij}$ variables to guarantee that they define a valid clustering on the nodes.
\begin{equation}
\label{eq:MainObj}
\MAXtwo{}{\sum_{i<j} w_{ij}^+(1-d_{ij}) + \sum_{i<j} w_{ij}^-d_{ij}\,,}
{d_{ij} \in \{0,1\},}
{ d_{ik} \leq d_{ij} + d_{jk} \text{ for all $i,j,k$. }} 
\end{equation}	
The first term counts the weight of agreements from clustering similar nodes together, and the second counts the weight of agreements from dissimilar nodes that are clustered apart.

For convenience, we encode the weights of a correlation clustering problem into a matrix $\mA$ by defining $A_{ij} = w_{ij}^+ - w_{ij}^-$. We think of $\mA$ as the 
adjacency matrix of a graph that has both positive and negative edges.
We can express the objective function in terms of $\mA$ as
\[ 
\MAX{}{ - \sum_{i<j} A_{ij}d_{ij} + \sum_{i<j} w_{ij}^+.}
\]
Since the second term is only a constant, to solve this problem optimally, we
can focus on just solving this ILP:
\begin{equation} 
\label{eq:MatrixObj1}
\MAXtwo{}{-\sum_{i<j} A_{ij}d_{ij}}
{d_{ij} \in \{0,1\},}
{ d_{ik} \leq d_{ij} + d_{jk} \text{ for all $i,j,k$. }}
\end{equation}
We can provide an alternative formulation of the correlation clustering
objective by introducing an indicator vector $\vx_i \in \{\ve_1, \ve_2, \ve_3,
\hdots , \ve_n \}$ for each node $i$, where $\ve_j$ is the $j^\text{th}$
standard basis vector in $\mathbb{R}^n$. This indicates which cluster node $i$ belongs to.
Unless each node ends up in its own singleton cluster, some of these basis
vectors will be unused.
We can then make the substitution $d_{ij} = 1 - \vx_i^T\vx_j$, since
$\vx_i^T\vx_j$ will be~$1$ if both nodes are in the same cluster but will
be~$0$ otherwise. After making this substitution and dropping a constant term in the objective, the problem becomes
\begin{equation} 
\label{eq:MatrixObj}
\MAXone{}{ \sum_{i<j} A_{ij}\vx_i^T \vx_j^{} }
{\vx_i \in  \{\ve_1,   \hdots , \ve_n \} \text{ for
		all $i = 1, \ldots , n$}\,.} 
\end{equation}
%

\section{Theoretical Results}

In this section, we present new results on the complexity of correlation
clustering under low-rank assumptions. In particular, we prove the problem
remains NP-hard when the underlying matrix has even one negative eigenvalue.
We are more concerned, however, with solving correlation clustering on
low-rank positive semidefinite adjacency matrices, in which case we give a polynomial
time solution. This scenario is analogous to results for other related optimization problems that admit polynomial-time solutions for low-rank positive semidefinite input matrices~\cite{ferrez2005solving,markopoulos2014optimal},
and is also a particularly natural assumption for the correlation clustering objective.
For example, if the adjacency matrix represents a correlation matrix (i.e., each entry is the Pearson correlation coefficient between two random variables), the input is already positive semidefinite.
So taking a low-rank approximation will yield the type input matrix studied here;
we give two examples of correlation clustering on a correlation matrix in Section~\ref{sec:applications}.

Even when the input is not a correlation matrix, we note that the correlation
clustering objective does not depend on the diagonal of the input matrix, so
we are able to shift the diagonal entries until the matrix is positive
definite before taking a low-rank approximation. Though the quality of the
approximation will vary depending on how much the diagonal needs to be
increased, this provides a means to apply our methods to get an approximate
solution for every full-rank dataset.

\subsection{Positive Semidefinite CC}

\paragraph{Rank-1 Positive Semidefinite} 
The simplest case to consider is when $\mA$ is rank-$1$ with one
positive eigenvalue. Because $\mA$ is symmetric, we can express it as $\mA =
\vv\vv^T$ for some $\vv \in \mathbb{R}^n$. In this case a perfect clustering
always exists and is easy to find: one cluster includes all nodes with
negative entries in $\vv$, while the other includes those with positive entries.
Nodes~$i$ and~$j$ are similar if and only if $A_{ij} > 0$, which is true if and only if
entries~$i$ and~$j$ of the vector~$\vv$ have the same sign.
So this simple two clustering agrees perfectly with the similarity labels.

In fact, the rank-$1$ positive semidefinite correlation clustering problem
is equivalent to maximizing a
quadratic form $\vx^T \mA \vx$ in binary variables ${\vx \in \{-1,1\}^n}$,
under the assumption that $\mA$ is rank-$1$.
This maximization problem can be solved in polynomial time for every
fixed low rank, $d$~\cite{markopoulos2014optimal}.
While this gives us a nice result for correlation clustering on
rank-$1$ matrices, it does not generalize to higher ranks as it only can
partition a graph into exactly \emph{two} clusters.

\paragraph{Rank-$d$ Positive Semidefinite} 
If the matrix $\mA$ is positive semidefinite but of rank $d > 1$,
there is no guarantee that a perfect partitioning exists, and the optimal clustering may have more than two clusters. We still begin by expressing $\mA$ in terms of low rank factors, i.e., $\mA = \mV\mV^T$ for
some $\mV \in \mathbb{R}^{n \times d}$.
Each node
in the signed graph can now be associated with one of the row vectors $\vv_1,
\vv_2, \hdots , \vv_n \in \mathbb{R}^{d \times 1}$ of $\mV$.
The similarity
scores between nodes~$i$ and~$j$ are given by $A_{ij} = \vv_i^T\vv_j^{}$, so
we can view this version of correlation clustering as the following vector partitioning
problem. Separate~$n$ points, or vectors, in $\mathbb{R}^d$ based on similarity scores given by dot products of the vectors:
\begin{theorem} \label{thm:cone} 
	If $\mA = \mV\mV^T$ for $\mV \in \mathbb{R}^{n \times d}$, then problem (\ref{eq:MatrixObj}) can be solved by partitioning the row vectors $\vv_1, \vv_2, \hdots \vv_n \in \mathbb{R}^{d\times 1}$ of $\mV$ into $d+1$ clusters $\{C_1, C_2,
	\hdots, C_{d+1}\}$ to solve
	\begin{equation}
	\label{eq:vec}
	\MAX{}{ \sum_{i=1}^{d+1} \norm{S_i}_2^2\,,}
	\end{equation}
	where we refer to the vector $S_i = \sum_{\vv \in C_i} \vv$ as the
	\textbf{sum point} of the $i^{\text{th}}$ cluster (in an empty cluster, defined to be
	the zero vector).
\end{theorem}

\proof We will show in two steps that when $\mA = \mV\mV^T$, the
clustering that maximizes objective~\eqref{eq:MatrixObj} also maximizes
objective~\eqref{eq:vec}.
The first step is to prove that~\eqref{eq:MatrixObj} is equivalent to maximizing the sum
of squared norms of sum points, where the maximization is taken over every
possible clustering. Second, we show that the objective function is maximized by a clustering with $d+1$ or fewer clusters.
\paragraph{Step 1: Maximizing a function on sum points} 

By doubling the objective function in (\ref{eq:MatrixObj}) and adding the constant
$\sum_{i=1}^n \vv_i^T\vv_i$, we obtain a related objective function that is maximized by the same clustering:
\begin{align}
\label{eq:HalfWay}
2 \sum_{i < j} \vv_i^T \vv_j (\vx_i^T \vx_j) + \sum_{i=1}^n
\vv_i^T\vv_i = \sum_{i = 1}^n \sum_{j=1}^n \vv_i^T \vv_j (\vx_i^T \vx_j)\,.
\end{align}
Since~$\vx_i$ and~$\vx_j$ are indicator vectors, identifying
which clusters nodes~$i$ and~$j$ belong to, the right-hand side of
equation~\eqref{eq:HalfWay} only
counts the product $\vv_i^T\vv_j^{}$ when $\vx_i^T \vx_j^{} = 1$.
So we are restricting our attention to
inner products between vectors that belong to the same cluster.
The contribution to the objective from cluster~$C_k$ is
\begin{align*}
\sum_{a \in C_k} \sum_{b \in C_k} \vv_{a}^T \vv_{b}^{} & = 
S_k^T S_k^{} = ||S_k||_2^2\,.
\end{align*}
Summing over all clusters completes step one of the proof.

\paragraph{Step 2: Bounding the number of clusters} 
To see that the number of clusters is bounded, observe that in the optimal clustering all the sum points must have pairwise non-positive dot products. Otherwise, there would exist distinct clusters~$C_i$ and~$C_j$ with $S_i^TS_j > 0$, and therefore
\[ (S_i + S_j)^T(S_i + S_j) = S_i^T S_i + 2S_i^TS_j + S_j^T S_j >
S_i^TS_i + S_j^T S_j\,.\]
Hence we could get a better clustering by combining~$C_i$
and~$C_j$.

Now, if there were an optimal clustering with two sum points that are orthogonal
-- $S_i^T S_j = 0$, we could combine the two clusters without changing the objective score.
Therefore, among all optimal clusterings, the one with the fewest clusters has the property that all sum points have pairwise negative dot products. The bound of~$d+1$ clusters then follows from the fact that the maximum number of vectors in $\mathbb{R}^d$ with pairwise negative inner products is $d+1$ (Lemma~8 of Rankin~\cite{Rankin}).
\hfill $\square$

It is worth noting that despite a significant difference in
motivation, our new objective function~\eqref{eq:vec} is nearly identical to
one used by Newman as a means to approximately solve maximum modularity
clustering~\cite{newman2006finding}. This is an interesting new connection between two clustering techniques that were not previously known to be related.
Newman and Zhang's work contains further
information on modularity \cite{zhang2015multiway,newman2006finding}.

The importance of Theorem~\ref{thm:cone} is that it
expresses the low-rank positive semidefinite correlation clustering problem as
a convex functional on sums of vectors in $\mathbb{R}^d$. Our problem is
therefore an instance of the well-studied vector partition
problem~\cite{onn2001vector,hwang1999polynomial}. Onn and Schulman showed that
for dimension~$d$ and a fixed number of clusters~$p$, this problem can be
solved in polynomial time by exploring the $O(n^{d(p-1) - 1})$ vertices of a $d(p-1)$-dimensional polytope called the \emph{signing zonotope}.

\begin{corollary}
	Correlation clustering with rank-$d$ positive semidefinite matrices (PSD-CC) is a special case of the vector partition problem with $d+1$ clusters, and is therefore solvable in polynomial time.
\end{corollary}

We later show how to construct a polynomial-time algorithm for
PSD-CC by reviewing the results of Onn and Schulman~\cite{onn2001vector}.
Before this, we observe a second theorem, which highlights an important geometric feature satisfied by the optimal clustering. It provides a first intuition as to how we can solve the problem in polynomial time.

\begin{theorem}
	\label{thm:ctwo}
	In the clustering $\textbf{C}_{opt}$, which
	maximizes~\eqref{eq:vec}, the $n$ row vectors of $\mV$ will be separated into
	distinct convex cones that intersect only at the origin. More precisely, if
	vectors $\vv_{x_1}, \vv_{x_2}, \hdots, \vv_{x_k}$ are all in the same cluster
	$C_x$ in $\textbf{C}_{opt}$, and $\vv_y \in \mathbb{R}^d$ is another row
	vector that satisfies $\vv_y = \sum_{i=1}^k c_i \vv_{x_i} \text{ for $c_i \in
		\mathbb{R}^{+}_0$ },$ then $\vv_y$ is also in cluster $C_x$.
\end{theorem}

\proof First notice that in $\textbf{C}_{opt}$, every vector $\vv$ must be more similar to its own sum point than to any sum point of a different cluster. To see this, assume that $\vv$ is in cluster $C_i$ with sum point $S_i$, but $\vv$ is more similar to another sum point $S_j$, i.e. 
\[\vv^TS_i < \vv^TS_j.\]
The contribution to the objective from the two sum points is $S_i^TS_i^{} + S_j^TS_j^{}$. If we move $\vv$ from cluster $C_i$ to cluster $C_j$, the contribution to the objective for the two new clusters is
\begin{align*}
&(S_i - \vv)^T(S_i - \vv) + (S_j + \vv)^T(S_j+\vv) = \\
&	S_i^TS_i^{} - 2\vv^TS_i +  \vv^T\vv + S_j^TS_j^{} + 2\vv^TS_j +  \vv^T\vv
\end{align*}
which is a higher score since $\vv^TS_i < \vv^TS_j$, contradicting the optimality of the first clustering. So in the optimal clustering every point is more similar to its own sum point than any other sum point.

Given this first observation we will now prove the main result of the theorem by contradiction. Assume that we have $k$ points $\vv_{x_1}, \vv_{x_2}, \hdots, \vv_{x_k}$ that in $\textbf{C}_{opt}$ are in cluster $C_x$ with sum point $S_x$. Let $\vv_y$ be another point in the dataset such that $\vv_y = \sum_{i=1}^k c_i \vv_{x_i}$ where $c_i > 0$ for $i = 1,2, \cdots , k$, and assume that $\vv_y$ is in a different cluster $C_y$ that has sum point $S_y$. By our first observation, every point in $C_x$ must be more similar to $S_x$ than $S_y$, so for $1\leq i \leq k$ we have that $\vv_{x_i}^TS_x > \vv_{x_i}^TS_y,$ which implies $c_i\vv_{x_i}^TS_x > c_i\vv_{x_i}^TS_y,$ for any positive scalar $c_i$. It follows that
\[ \vv_y^TS_x = \sum_{i = 1}^k c_i\vv_{x_i}^TS_x > \sum_{i = 1}^k c_i\vv_{x_i}^TS_y = \vv_y^TS_y. \]
This indicates that $\vv_y$ is more similar to $S_x$ than to the sum point of the cluster to which is belongs, which is a contradiction.

\hfill $\square$

\begin{figure}
	\centering
	\includegraphics[width=.45\linewidth]{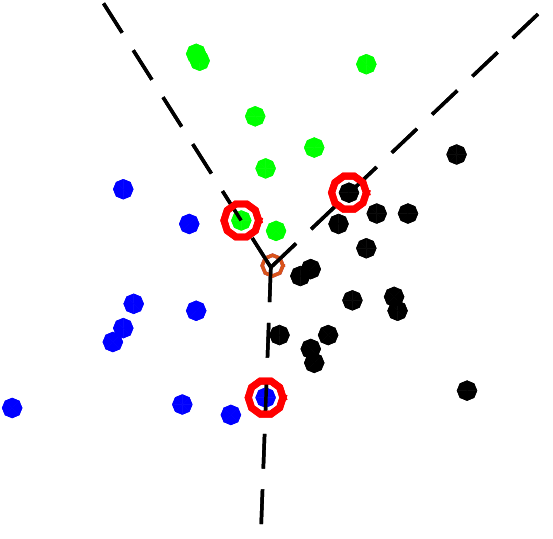}
	\caption{The three-clustering that solves a small rank-2 correlation clustering problem. Each cluster is shown by a different color, and can be delimited on each side by rays through the origin. By selecting the right combination of three points (circled in red), we will be able to find the optimal clustering.}
	\label{fig:r2cc}
\end{figure}

Theorem \ref{thm:ctwo} implies that we can find an optimal clustering
in polynomial time by checking every possible partitioning of the vectors into
convex cones. By Theorem 2.7 of Klee~\cite{klee1955separation}, every pair of cones
can be separated by a hyperplane through the origin.
Furthermore, Cover \cite{cover1967geometrical} proved that for every set of~$n$ points in
$\mathbb{R}^d$, there are $O(n^{d-1})$ such hyperplanes that split the points into two groups. This means that to cluster the points into $d+1$ convex cones, we must choose
${d+1 \choose 2}$ of the $O(n^{d-1})$ hyperplanes so that each distinct pair
of clusters is separated by a hyperplane. This gives us a total of $O(n^{(d-1)
	{d+1 \choose 2}})$ ways to cluster the points into $d+1$ convex cones, which
we can enumerate in polynomial time. When $d=2$, we can
efficiently find the boundaries of the optimal convex cones by considering
the~$n$ rays that each connect the origin to one of the~$n$ points.
Since there are at most three clusters in this case, we can solve the problem by testing
$O(n^3)$ triplets of points as possible separators for the clusters, each time evaluating the objective.
A visualization of this process is shown in Figure~\ref{fig:r2cc}.
Though we can make this procedure very efficient for the two-dimensional case, it is significantly less efficient for dimension~$d$ greater than~$2$, so we resort to the methods proposed by Onn and Schulman~\cite{onn2001vector}, which we review in Section~\ref{sec:algorithms}.

\subsection{Rank-$1$ Negative Semidefinite}

When $\mA$ has rank~$1$ and its nonzero eigenvalue is negative, we know $\mA =
-\vv\vv^T$ for some $\vv \in \mathbb{R}^d$. This changes the objective~\eqref{eq:vec}
in three ways: the negative sign converts the maximization to a minimization, the entries of $\vv$ are real numbers rather than row vectors, and the upper bound of~$d+1$ clusters no longer applies. Solving the Rank One Negative Eigenvalue Correlation Clustering problem (RONE-CC) is therefore equivalent to
\begin{equation}
\label{eq:RONE}
\min_{C,k} \hspace{.2cm} \sum_{i =1}^k \Big(\sum_{v \in C_i} v \Big)^2\, .
\end{equation}

The following theorem regarding RONE-CC is analogous to a result of
Papailiopoulos et al., who use a reduction from Subset Sum to prove that the Densest-$k$-Subgraph problem is NP-hard for input matrices of rank-1 with one negative eigenvalue~\cite{papailiopoulos2014finding}. Our result relies on a reduction from the related Partition problem, and accounts for the fact that in an instance of correlation clustering we must optimize over the set of partitionings with arbitrarily many clusters.

\begin{theorem}
	\label{thm:rone}
	Rank-one Negative Eigenvalue Correlation Clustering is NP-Complete.
\end{theorem}
\proof Given that general correlation clustering is NP-Complete, we know that the decision version of RONE-CC must also be in NP. To show the problem is NP-hard, we can use a reduction from the Partition problem, one of the NP-hard problems listed by Garey and Johnson~\cite{gary1979computers}. For this problem we are given a multiset of $n$ positive integers and seek to partition the set into subsets of equal sum. 

Consider a multiset of $n$ positive integers. Let $s$ be the smallest integer and $B$ be the sum of all $n$ integers. We assume $s$ and $B$ are both even--if this is not the case we can multiply all numbers in the set by two so that this assumption is satisfied. Letting $M = B/2 - s/2$, we add two copies of~$-M$ to the input: now the total sum of the input integers is~$s$. We can show that the optimal solution to RONE-CC on this input will perfectly partition the $n$ positive integers into two subsets of equal sum if such a perfect partition exists.

Assume the positive integers can be split into two subsets of sum $B/2$.
If we include exactly one copy of the $-M$ values with each of these two subsets, then each cluster sums to $B/2 - M = s/2$. The RONE-CC objective corresponding to this two-clustering is $2\left(s/2 \right)^2 = s^2/2$.
We now show that every other clustering yields a worse objective value.
Clearly, any clustering with one copy of~$-M$ in each cluster that does not equally split the positives
will have objective score~$> s^2/2$. If we cluster all integers together, the sum is~$s$, and the objective would be $s^2 > {s^2}/{2}$. If on the other hand we consider a two-clustering where both $-M$ values are in the same partition, or any clustering with more than two clusters,
then there must exist some cluster with only positive integers. This cluster has sum at least $s$, leading to an objective of at least~$s^2$. The best option is therefore to form two clusters, each of which contains one of the $-M$ values and a subset of the $n$ positive integers summing to $B/2$.
\hfill$\square$

\section{Algorithms}
\label{sec:algs}
\label{sec:algorithms}
In this section we show how to obtain a polynomial-time algorithm for
solving PSD-CC. We first review the results of Onn and
Schulman~\cite{onn2001vector}, which establish the existence of a
polynomial-time algorithm, by analyzing the properties of a $d^2$-dimensional polytope
called the signing zonotope. We then combine this with a vertex-enumeration
procedure developed by Stinson, Gleich, and Constantine~\cite{stinson2016randomized}.

\subsection{Signing Zonotope}
A zonotope is the linear projection of a high-dimensional hypercube
into a lower-dimensional vector space. We are primarily concerned with the
\emph{signing zonotope} introduced by Onn and Schulman~\cite{onn2001vector}, whose vertices directly correspond to clusterings of the $n$ vectors of a vector partition problem.

Consider a set $S_V$ of $n$ vectors $\vv_1, \vv_2, \hdots , \vv_n \in \mathbb{R}^{d \times 1}$ in an instance of the vector partition problem. A \emph{signing} of these vectors is defined to be a vector $\sigma = (\sigma_{a,b}^i)\in
\{-1,1\}^M$, where $M = n {d+1 \choose 2}$. Each entry $\sigma_{a,b}^i$ uniquely corresponds to a triplet $(\vv_i, a,b)$, where $\vv_i$ is one of the data points we are clustering and $1 \leq a < b \leq (d+1)$ are the indices for two distinct clusters in a $(d+1)$-clustering of the $n$ vectors. If $b < a$, we define $\sigma_{a,b}^{i} = -\sigma_{b,a}^i$ and
associate each signing with a matrix $\mT_\sigma$:
\begin{equation}
\label{eq:Tsig}
\mT_\sigma = \sum_{i=1}^n\, \sum_{1 \leq a < b \leq d+1}
\sigma_{a,b}^{i} \vv_i \cdot (\textbf{e}_a -\textbf{e}_b)^T \in \mathbb{R}^{d
	\times (d+1)} \,,
\end{equation}
where $\ve_a,\ve_b \in \mathbb{R}^{(d+1) \times 1}$ are the
$a^{\text{th}}$ and $b^{\text{th}}$ standard basis vector, respectively.
By construction, the row sum of $\mT_\sigma$ will be the zero vector, so if we
are given the first $d$ columns of the matrix we will be able to recover the
last column even if it is not given explicitly. We associate with each signing
$\sigma$ a vector $Z_\sigma$ of length $d^2$ made by stacking the first $d$
columns of $\mT_\sigma$ on top of one another. Here we will refer to this
vector as the Z-vector of~$\sigma$. A signing~$\sigma$ is said to be \emph{extremal} if its Z-vector is a vertex of the signing zonotope, which we define below. Furthermore, Onn and Schulman proved that for every vertex~$v$ of the zonotope, there exists exactly one extremal
signing~$\sigma$ such that~$v$ is the Z-vector of~$\sigma$. In other words, the extremal signings are in one-to-one correspondence with the vertices of the zonotope.

The \emph{signing zonotope}~$\mathcal{Z}$ for this instance of the vector partition problem is defined to be
\[ \mathcal{Z} = \text{conv} \{Z_\sigma: \sigma \text{ is a signing of
	$S_V$} \}\,.\]
In other words,~$\mathcal{Z}$ is the convex hull of all~$2^M$ Z-vectors of
signings of~$S_V$. 
We now state two important results established by Onn and Schulman~\cite{onn2001vector} about the signing zonotope.

\begin{theorem}
	(Results from \cite{onn2001vector}) The following properties hold regarding the signing zonotope~$\mathcal{Z}$ introduced above:
	
	\begin{enumerate}
		\item Each vertex of~$\mathcal{Z}$ can be mapped to a
		clustering of the~$n$ vectors in~$S_V$, where each cluster is contained in one of $d+1$ convex cones. Additionally, there exists an extremal signing that maps to the clustering which optimizes the objective of the vector partition problem.
		\item Signing zonotope~$\mathcal{Z}$ has at most $O(n^{d^2 - 1})$ vertices.
	\end{enumerate}
\end{theorem}

All we need then to solve the vector partition problem, and hence
PSD-CC, is to iterate through each extremal signing of~$\mathcal{Z}$, obtain
the clustering it corresponds to, and evaluate objective~\eqref{eq:vec} for that clustering. At the end we output the clustering with the maximum objective value.

The procedure for associating an extremal signing with a clustering of $S_V$ is given in Proposition~2.3
of Onn and Schulman's work~\cite{onn2001vector}.
This states that for all $i = 1, 2, \hdots , n$, there exists a unique index $1 \leq c_i \leq d+1$ such that $\sigma_{c_i, k}^i = 1$ for all $k \neq c_i.$ Thus vector $\vv_i$ belongs to cluster number $c_i$ in the optimal clustering.

We can interpret this fact in light of Theorem~\ref{thm:ctwo}.
Recall that according to Klee~\cite{klee1955separation}, between each pair of
clusters $C_a$ and $C_b$ there exists a hyperplane that separates $\mathbb{R}^d$ into two half spaces such that $C_a$ is
in one half and $C_b$ is in the other.
Each extremal signing $\sigma$ encodes information about which side of the separating hyperplane each vector $\vv_i$ is on. For example, $\sigma_{a,b}^i  = 1$ indicates that vector $\vv_i$ is the same half space as all vectors in cluster $C_a$. More importantly, this tells us that $\vv_i$ is \emph{not} in the same half space as cluster $C_b$, so we rule out the possibility that $\vv_i$ is in cluster $C_b$. On the other hand, $\sigma_{a,b}^i  = -1$ indicates that $\vv_i$ is on the same side as cluster $C_b$, eliminating the possibility that $\vv_i$ is in $C_a$. If we consider all entries of $\sigma$, Onn and Schulman's proposition effectively tells us that there is only one cluster that will not be ruled out by this process, so by default this is the cluster where $\vv_i$ is located. From this elimination process we recover a $(d+1)$-clustering of the vectors. With this, we are now able to state the exact runtime for solving PSD-CC.	
\begin{theorem}
	The fixed-rank positive semidefinite correlation clustering problem can be solved in $O(n^{d^2})$
	time.
\end{theorem}	
\proof	Relying on previous complexity and algorithmic results regarding
zonotopes, in their Corollary~3.3, Onn and Schulman
establish that the $(d+1)$-vector partition problem can be solved with
$O(n^{d^2 -1})$ operations and queries to an oracle for evaluating the convex
objective functional~\cite{onn2001vector}.
We now show that it takes $O(n)$ operations to evaluate our specific oracle function
for each of the $O(n^{d^2-1})$ extremal signings. In our case, the complexity
of the oracle is the time it takes to evaluate summation~\eqref{eq:vec} for a
given extremal signing~$\sigma$.
Treating~$d$ as a fixed constant, this procedure involves inspecting the $M = O(n)$ entries of~$\sigma$ to identify a clustering, and~$O(n)$ operations to add vectors in each cluster to obtain the sum points.
We require only a constant number of operations to take dot products of the sum points
and add the results, so the evaluation process takes~$O(n)$ time, and hence
the overall process $O(n^{d^2})$ time.

	\subsection{Practical Algorithm}
	
	Though theoretically polynomial time,
	the runtime given above is impractical for applications.
	We turn our attention to an algorithm which approximately solves the PSD-CC objective, but is much more efficient in practice.
	We implement a randomized algorithm for sampling vertices of a zonotope (that will eventually enumerate them all), developed by
	Stinson, Gleich, and Constantine~\cite{stinson2016randomized}. 
	When mapping a hypercube in $\mathbb{R}^M$ onto a zonotope in
	$\mathbb{R}^N$, the basic outline of their procedure is as follows.
	Form an $N \times M$ matrix $\mG$, where each column is a generator of the zonotope, i.e.,
	$\mG$ is the linear map that maps the hypercube into a lower-dimensional space.
	Given a vector $\textbf{x}$ drawn from a standard Gaussian distribution,
	compute $\textbf{v} = \mG\sign(\mG^T\textbf{x})$, where $\sign(\textbf{u})$
	returns a vector with $\pm 1$ entries, reflecting the signs of the entries of
	$\textbf{u}$. The main insight of Stinson, Gleich and Constantine is that under reasonable assumptions on $\mG$, $\textbf{v}$ will be
	a vertex of the zonotope. One can construct the entire zonotope by generating vertices in this way, by
	checking whether a given vertex has been previously found, and continuing
	until all the vertices have been returned. In practice, it is better to just approximate the zonotope by stopping after a certain number of vertices have been found. 
	
	We alter this procedure slightly to fit our needs. Note
	that the generators of the signing zonotope come from outer products of
	the form $\vv_i\cdot (\textbf{e}_r -\textbf{e}_s)^T$, for $i = 1, 2, \hdots, n$
	and $1 \leq r < s \leq d+1$. This product gives a $d \times (d+1)$ matrix with
	a zero row sum, so taking the first $d$ columns and stacking them into a
	vector we get one of the columns of $\mG$. Equation~\eqref{eq:Tsig} shows that when we form a linear combination
	of these generators, where the coefficients of the linear combination are entries of a signing $\sigma$, the output is exactly
	the Z-vector corresponding to $\sigma$. 
	We are ultimately interested in
	extremal signings rather than actual zonotope vertices, so we repeatedly
	generate vectors $\sigma = \sign(\mG^T\textbf{x})$.
	We then inspect the
	entries of~$\sigma$ and find the corresponding clustering of~$n$ vectors and
	thence the clustering's PSD-CC objective score~\eqref{eq:vec}.
	We do this for a very large number of
	randomly generated extremal signings and output the one with the highest
	score.  We name our algorithm based on this zonotope vertex enumeration, \alg{ZonoCC},
	outlined in Algorithm~\ref{alg:ZonoCC}.
	\begin{algorithm}[h]
		\caption{\alg{ZonoCC}}
		\label{alg:ZonoCC}
		\begin{algorithmic}
			\STATE \textbf{Input:} rows of $\mV$: $\vv_1, \vv_2, \hdots , \vv_n \in \mathbb{R}^d$, and $k \in \mathbb{N}$
			\STATE Form generator matrix $\mG$
			\STATE Set $\text{BestClustering} \gets \varnothing$,
			$\text{BestObjective} \gets 0$
			\FOR{$i = 1,2, \hdots, k$}
			\STATE 1. Generate standard Gaussian $\vx \in \mathbb{R}^M$
			\STATE 2. $\sigma \gets \sign(\mG^T\vx)$
			\STATE 3. Determine clustering $C$ and objective
			$C_{\text{obj}}$ from $\sigma$
			\IF{$C_{\text{obj}} > \text{BestObjective}$}
			\STATE Set $\text{BestClustering} \gets C$,
			$\text{BestObjective} \gets C_{\text{obj}}$
			\ENDIF
			\ENDFOR
			\STATE \textbf{Output:} $\text{BestClustering}, \text{BestObjective}$
		\end{algorithmic}
	\end{algorithm}	
	
	Our new algorithm is significantly faster than exploring all vertices of the zonotope. Once we have formed the $n {d+1 \choose 2}$ columns of~$\mG$, generating~$\sigma$ by matrix multiplication and determining the corresponding clustering both take~$O(n)$ time since~$d$ is a
	small fixed constant. The overall runtime is therefore just~$O(nk)$, where~$k$ is the number of iterations. Although \alg{ZonoCC} is not guaranteed to return the optimal clustering, Stinson, Gleich, and Constantine prove that with high probability the zonotope vertices that are generated will tend to be those  which most
	affect the overall shape of the zonotope~\cite{stinson2016randomized}. We expect such extremal vertices of the zonotope to be associated with extremal clusterings of the~$n$ data points, i.e., clusterings with high objective score.

\section{Numerical Experiments}
\label{sec:applications}

In this section, we demonstrate the performance of \alg{Zono-CC} in a variety
of clustering applications.
In order to understand how its behavior depends on the rank,
as well as the size of the problem,
we begin by illustrating the performance of \alg{Zono-CC}
on synthetic datasets that are low-dimensional by construction.
We then study correlation clustering in two real-world scenarios:  (i) the volume of search queries over time for computer science conferences and (ii) stock market closing prices for S\&P 500 companies.
Since neither of these cases is intrinsically low-rank,
we study the performance of \alg{Zono-CC} on low-rank approximations of the data.
Curiously, the best results are achieved on extremely
low-rank approximations.
Our goal in both the synthetic and real-world experiments is to compare our algorithm to other well-known correlation clustering algorithms and, when
possible, see how well our algorithm is able to approximate the optimal solution.
On the real-world data, we also run the \alg{$k$-means} procedure
and find that it is unable to create the clustering we find via correlation clustering. 

For our last experiment, we show how to cluster any unsigned network with
\alg{ZonoCC}, by first obtaining an embedding of the network's vertices into a
low-dimensional space. We use this technique to cluster and study the
structure of several networks from the Facebook 100
dataset~\cite{Traud-2012-facebook}. Here we compare
\alg{ZonoCC} against \alg{$k$-means}, an algorithm that is very commonly applied to
cluster data in low-dimensional vector spaces. The goal of this final
experiment is not to show that \alg{ZonoCC} is better in itself, but to
analyze the results of \alg{ZonoCC} when both high-quality 
and low-quality embeddings of the vertices are applied.
We find that with lower-quality embeddings, 
\alg{ZonoCC} is able to better uncover meaningful structure in the networks than \alg{$k$-means}.

Our experiments revolve around the following four algorithms, three of which are specifically intended for correlation clustering. In our experiments we show runtimes for guidance only and note that these are non-optimized implementations. We make code for all of our algorithms and experiments available at \texttt{https://github.com/nveldt/PSDCC}.

\paragraph{\textbf{\alg{Exact ILP}}} 
\noindent
For small problems, we compute the optimal solution to the correlation clustering problem by solving an integer linear program with the commercial software Gurobi. 

\paragraph{\textbf{\alg{CGW}}}
\noindent
The $0.7664$-approximation for maximizing
agreements in weighted graphs, based on a semidefinite programming
relaxation, by Charikar, Guruswami, and Wirth~\cite{charikar2005clustering}.

\paragraph{\textbf{\alg{Pivot}}}
\noindent
The fast algorithm developed by Ailon, Charikar,
and Newman for $\pm1$ correlation clustering
instances~\cite{ailon2008aggregating}. It uniformly randomly selects a vertex, clusters it with all nodes similar to it, and repeats.
Depending on problem size,
we return the best result from~1000 or~2000 instantiations.

\paragraph{\textbf{\alg{$k$-means}}}
\noindent
The standard Lloyd's $k$-means algorithm, as implemented
in MATLAB, with $k$-means++ initialization~\cite{arthur2007k}.
Because of its speed, we return the best of 100 instantiations.

\subsection{Synthetic Datasets}

We begin by demonstrating that \alg{ZonoCC} computes a very good approximation to the optimal PSD-CC objective even though it does not test all vertices of the zonotope. We use several synthetic datasets for a range of $d$ values. We consider both the scenario where there is a true planted clustering in the dataset, as well as the case where there is no clear clustering stucture in the data. The first of these cases tests eachs algorithm's ability to detect a clustering when there is a high signal to noise ratio. In the second experiment we are purely testing how well each algorithm can optimize the objective in the absence of any particular clustering structure in the dataset. We also compare the performance of \alg{Pivot} and \alg{ZonoCC} on larger datasets in a third experiment. Finally, in experiment four we demonstrate how \alg{ZonoCC} performs for a varying number of iterations.

\begin{figure}[tb]
	\centering
	\subfigure{%
		\includegraphics[width=1.75in]{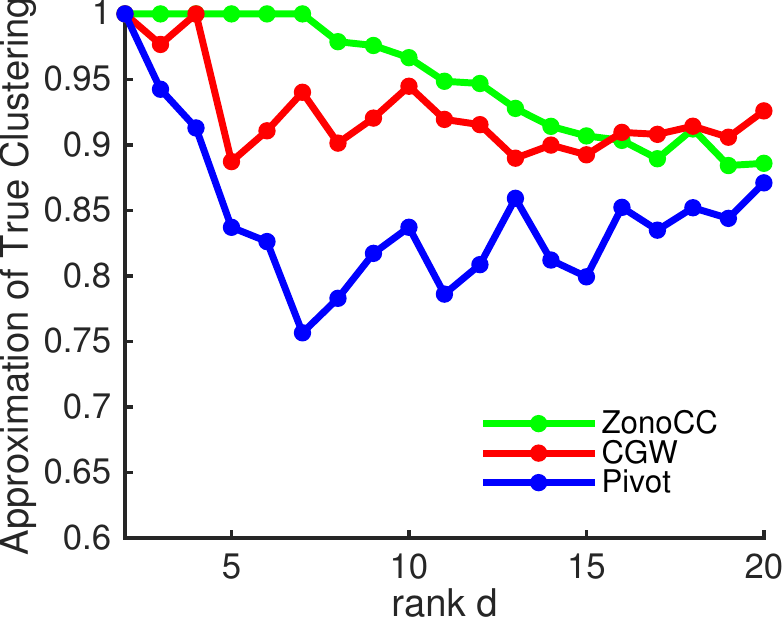}
		\label{fig:synplant1}} $\qquad$
	\subfigure{%
		\includegraphics[width=1.75in]{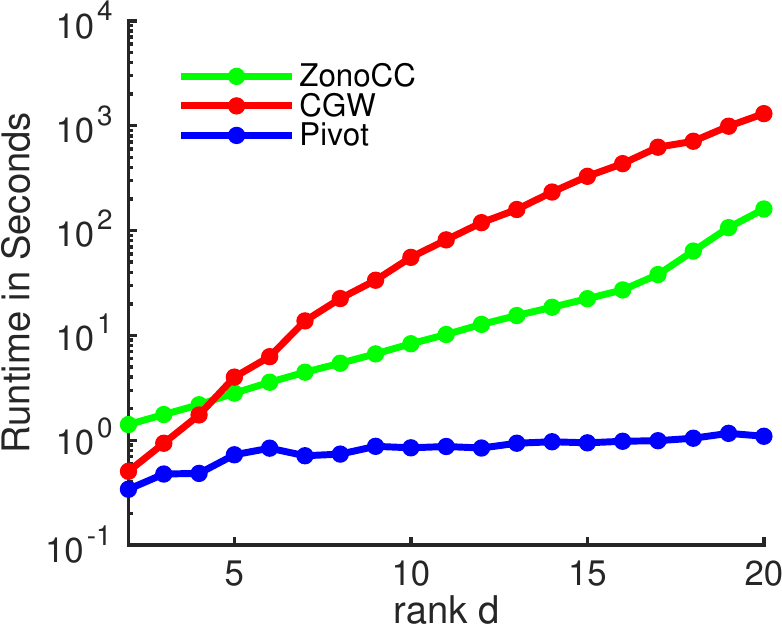}
		\label{fig:synplant2}}
	\caption{Results for \alg{ZonoCC} (green), \alg{CGW} (red), and \alg{Pivot} (blue) on synthetic datasets with a true underlying clustering structure. The left plot gives each algorithm's approximation to the score of the planted clustering for $d$ ranging from 2 to 20 and $n = 10d$. The right plot shows runtimes for each method. For both plots we take the median over five trials. \alg{ZonoCC} outperforms other algorithms for all values of $d$ up to 15. At this point \alg{CGW} finds a better clustering score, but does so at the expense of a much longer runtime. Even for high values of rank, we note that \alg{ZonoCC} achieves a score that is 85\% of the planted clustering.}
	\label{fig:synplant}
\end{figure}

For the first experiment, we perform the following steps to generate a dataset with a planted partitioning:
\begin{enumerate}
	\item For each rank $d$, we set $n = 10d$ and choose an integer $k$ between $d/2$ and $d+1$ uniformly at random to be the number of clusters to form.
	\item Assign each of the $n$ objects to one of the $k$ planted clusters uniformly at random.
	\item Assign to each of the planted clusters a vertex of a regular $d$-simplex, whose $d+1$ vertices in $\mathbb{R}^d$ all have pairwise negative dot products. 
	\item Form $\mW \in \mathbb{R}^{n\times d}$ by setting the $i$th row to be the coordinates of the simplex vertex assigned to the cluster that object $i$ belongs to. Now $\mW$ defines a correlation clustering problem with a perfect partitioning.
	\item Add noise to form matrix $\mV$ in the following way:
	\[ \mV = (1-\varepsilon) \mW + \varepsilon \mE \]
	where $\mE$ is an $n \times d$ matrix of standard Gaussian noise. For our experiments we choose $\varepsilon = .15$, which is enough noise to guarantee there will no longer be a perfect clustering, but so that the structure of the planted clustering is still detectable.
\end{enumerate}
After performing the above steps, computing $\mA = \mV\mV^T$ gives a synthetic dataset with an underlying clustering which will correspond to the optimal clustering or at least approximate it very well. We then run each correlation clustering algorithm on this dataset and compare each algorithm's objective score against the score of the planted clustering. This setup allows us to test how well each method is able to perform in the high signal to noise regime. In Figure \ref{fig:synplant} we display objective scores and runtimes for different algorithms on synthetic datasets with planted clusters for a range of $d$ values from 2 to 20. We use 50,000 itertions of \alg{ZonoCC} and 1000 instantiations of \alg{Pivot} in each case.

We note that for this first experiment \alg{ZonoCC} outperforms all other methods for matrices up to rank 15. At this point \alg{CGW} begins to take over in objective score, although it does so at the expense of a much longer runtime. Recall that we are running \alg{ZonoCC} for a fixed number of iterations; we would expect to see improved objective scores for running the algorithm longer as problem size increases. Note that for even for rank 20, \alg{ZonoCC} still achieves an objective score that is within 85\% of the score of the planted clustering.

\begin{figure}
	\centering
	\includegraphics[width=.4\linewidth]{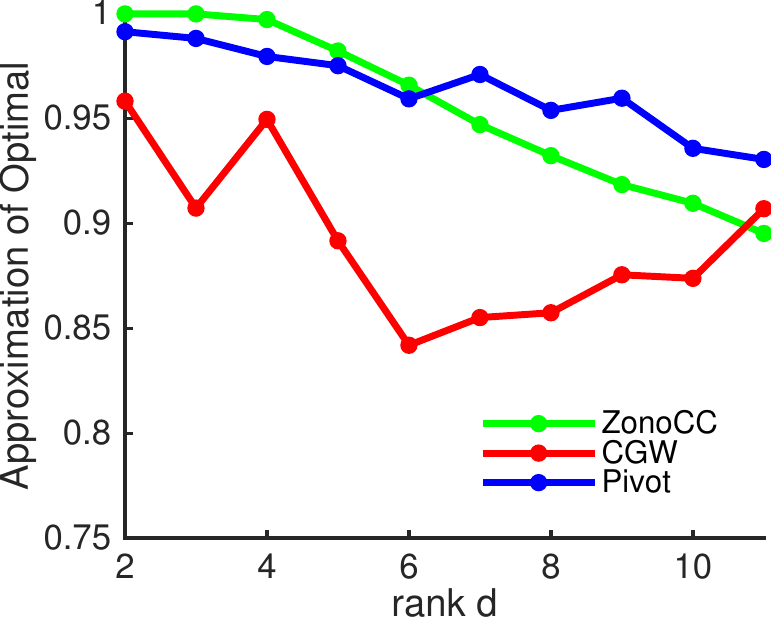}
	\caption{Median approximation ratios for \alg{ZonoCC} (green),
		\alg{Pivot} (blue), and \alg{CGW} (red) for synthetic datasets generated with no underlying clustering structure. We fix $n = 60$, vary the rank $d$, and take the median score over five trials. For low values of $d$, \alg{ZonoCC} outperforms the other
		correlation clustering methods, though as~$d$ increases, \alg{Pivot} and then
		\alg{CGW} take over.}
	\label{fig:syn}
\end{figure}

For the second experiment we generate synthetic datasets without any underlying structure by forming random $n \times d$ matrices $\mV$ with entries taken from a standard Gaussian distribution. We fix $n=60$, which is small enough so that we can obtain the exact solution to the correlation clustering problem for the matrix $\mA = \mV\mV^T$ by solving the ILP, though at a large computational expense. In this case we are purely testing each algorithm's ability to optimize the correlation clustering objective function, as there is no true clustering structure to detect. Figure~\ref{fig:syn} gives a visualization of the approximation ratio for \alg{ZonoCC} (with 50,000 iterations), \alg{CGW}, and 1000 instantiations of \alg{Pivot}. We give representative runtimes for each method in Table \ref{tab:run}. Although we expect this to be a worst-case scenario for our algorithm, we notice that \alg{ZonoCC} still outperforms the other methods for very low-values of rank. For these experiments \alg{Pivot} does extremely well, which we expect is because there are essentially many clusterings that acheive a good score. Just as before, even for relatively high values of rank, \alg{ZonoCC} is still able to find a clustering that is at least 85\% of optimal.

\begin{figure}[tb]
	\centering
	\subfigure{%
		\includegraphics[width=1.5in]{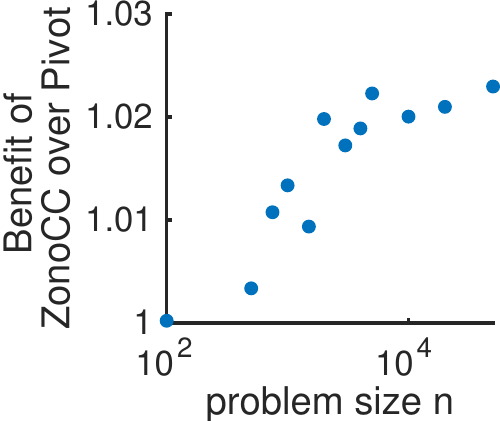}
		\label{fig:zvp1}} $\qquad$
	\subfigure{%
		\includegraphics[width=1.5in]{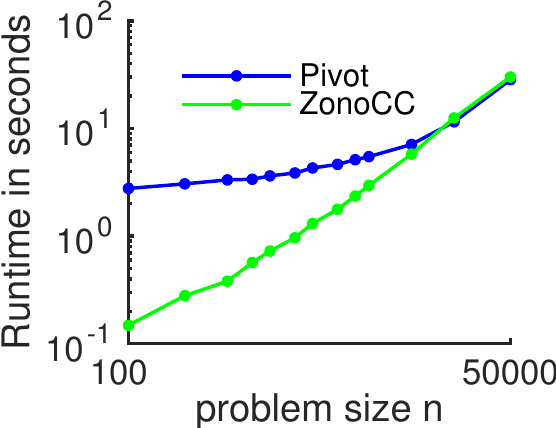}
		\label{fig:zvp2}}
	\caption{In the left we show the benefit of \alg{ZonoCC} over
		\alg{Pivot} in terms of objective score on synthetic datasets of increasing
		size $n$ and rank $d =5$. \alg{ZonoCC} always has a higher objective value,
		and as problem size grows, so does its performance over \alg{Pivot}. On the right we plot corresponding runtimes. \alg{ZonoCC} is faster for problem sizes under $n = 10000$. For higher values, times are comparable and scale roughly linearly in $n$.}
	\label{fig:scaling}
\end{figure}

In the third experiment, we wish to understand how \alg{ZonoCC} compares with \alg{Pivot} as we scale the problems up in \emph{size}, for a fixed rank $d=5$.
This setting renders both the ILP and \alg{CGW} methods infeasible,
and so we show results only for \alg{ZonoCC} with 1000 iterations, and 2000 instantiations of \alg{Pivot} in Figure~\ref{fig:scaling}.
These figures show us that \alg{ZonoCC} always outperforms \alg{Pivot} in objective, with the
ratio of objective scores slowly growing as the problem size increases.
We remark that the running time for each algorithm involves both generating clusterings as well as checking the objective value for each clustering. For small values of~$n$, computing the objective scores becomes a more noticeable fraction of the computation time. For this reason the runtime of \alg{ZonoCC} is much faster than \alg{Pivot} for small~$n$ because we are checking half the number of clusterings. As problem size increases, we see that both algorithms roughly scale linearly in~$n$.

\begin{figure}
	\centering
	\includegraphics[width=.4\linewidth]{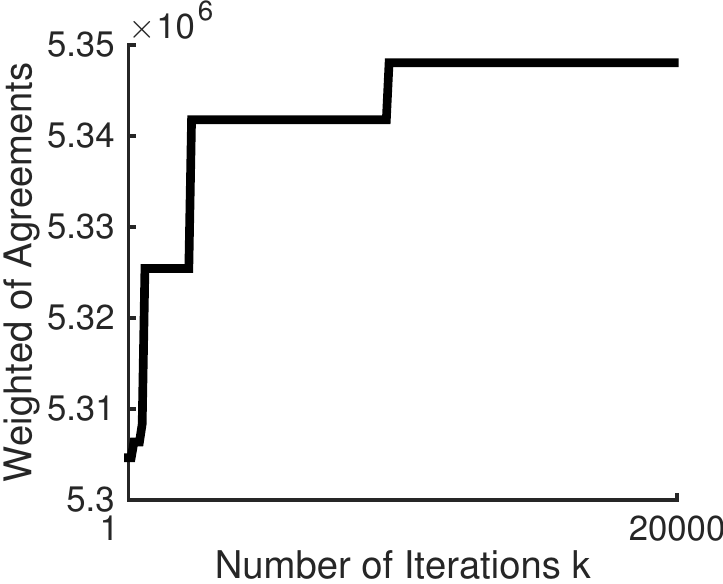}
	\caption{Best objective value as a function of number of iterations
		of~\alg{ZonoCC} for a synthetic dataset with $n = 3000$ and $d = 5$.
		\alg{ZonoCC} quickly finds a good-quality clustering, and slowly improves as we let the algorithm run longer.}
	\label{fig:iterations}
\end{figure}

In the final synthetic experiment, we study how the approximation changes as the number of iterations of \alg{ZonoCC} increases.
The results in Figure~\ref{fig:iterations} show that the algorithm quickly
attains a near-optimal solution, but moves closer to optimality slowly, as it
continues to explore more vertices of the zonotope. 
The results displayed are for $n = 3000$ and $d=5$, though this behavior is typical for most instances.


\begin{table}[t]
	\centering
	\caption{Median running time in seconds for Experiment 2, where we run each correlation clustering method on a dataset with no underlying clustering structure. We set $n = 60$ and test a range of values for the rank $d$.}
	\label{tab:run}
	\vspace{.2cm}
	\begin{tabular}{r r r r r}
		\toprule
		$d$ & Exact ILP & \alg{CGW} & \alg{ZonoCC} & \alg{Pivot}\\
		\midrule
		2	& 2	& 7	& 2	& 2\\
		3	& 8	& 8	& 3	& 2\\ 
		4	&24	&7&	3& 2\\
		5	&50	&6&4&2	\\
		6	&46	&6&	4&2	\\
		8	&1060&7&7&2	\\
		10	&1462&6&7&2	\\
		\bottomrule
	\end{tabular}
\end{table}

\subsection{Clustering Using Search Query Data}
\label{sec:search}
\label{sec:volume}

For our first real-world application we use \alg{ZonoCC} to cluster top-tier
computer science conferences based on search query volume data.
Each search
term is either a conference acronym (e.g., ``ICML''),
or is an acronym concatenated with ``conference'' (e.g., ``WWW conference'').
For each search term, we obtain from Google Trends
a time series of the volume of search queries
for each month over the course of a six-year period, 2010--16.
This data appears to fluctuate for repeated API calls,
so before clustering we smooth out the data to capture the overall trend
in each time series (using exponential smoothing with parameter $\alpha=0.5$).
Second, we remove the trend across all time series by fitting a quadratic
polynomial to the mean volume. Finally, we `$z$-score' normalize the time series.
We then calculate the correlation coefficient between each pair of conferences
using the processed data. This gives us a full-rank matrix of correlation
values, of which we take a low-rank approximation to feed to \alg{ZonoCC}.

Using the rank-$3$ approximation gives us the optimal clustering (as determined by the ILP).
In this case, both \alg{Pivot} and \alg{CGW} also find the optimal solution.
We show the runtime and objective values in the upper part of Table~\ref{tab:cs}.  The optimal clustering consists of three clusters: the two large clusters and an outlier set of only one conference, the International Conference on Computer Graphics Theory and Application.

\begin{table}[h]
	\centering
	\caption{Objective scores and runtimes in seconds for correlation clustering
		algorithms on two real-world datasets. Due to the size of the stocks dataset, we can run only \alg{ZonoCC} and
		\alg{Pivot} on it.}
	\label{tab:cs}
	\begin{tabularx}{\linewidth}{llXXXX}
		\toprule
		Dataset & & \alg{ZonoCC} &\alg{Pivot} & \alg{CGW} & \alg{ILP} \\
		\midrule 
		CS Conf. & Obj. & 7540.0 & 7540.0 & 7540.0 & 7540.0 \\
		$n=157$ & Time &  7 & 1 & 1380& 52 \\
		\midrule
		Stocks & Obj. & 5100.2 & 5099.5 & --- & --- \\
		$n=497$ & Time & 40 &  20& --- & --- \\
		\bottomrule 
	\end{tabularx}
\end{table}

\begin{figure}[tb]
	\centering
	\subfigure{%
		\includegraphics[width=2in]{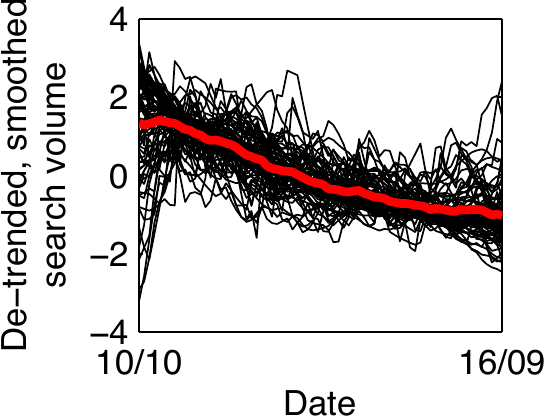}
		\label{fig:search1}} $\qquad$
	\subfigure{%
		\includegraphics[width=2in]{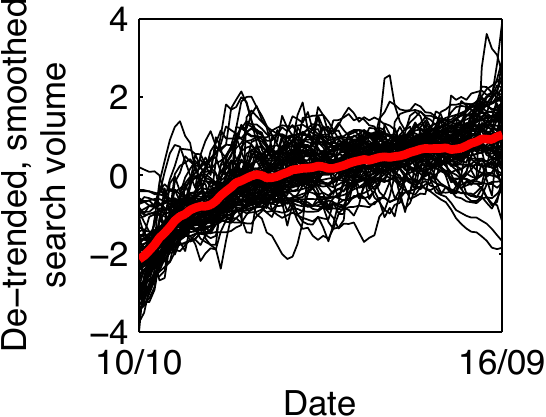}
		\label{fig:search2}}
	\caption{Smoothed search query data for top-tier computer
		science conferences. The partition discovered by \alg{ZonoCC} splits the conferences into those whose search volume shows an overall decreasing trend (left), and an increasing trend (right); WWW is in the increasing cluster. For each plot we show an average of the clusters as a thick red line. There is also a third cluster with only one outlier conference that is not shown here.}
	\label{fig:cs}
\end{figure}

{It is interesting to observe the significance of the optimal clustering of this dataset.}
In Figure~\ref{fig:cs}
we plot for each cluster the smoothed search query data for all of the conferences in the cluster.
We observe that \alg{ZonoCC} effectively partitions the dataset into conferences that have increased in search query volume over the course of the past six years, and those that have experienced an overall decrease in search volume. We expect it to be unsurprising to our readers that the WWW conference is in the ``growing'' cluster. We are unable to find any set of three clusters from \alg{$k$-means} that
resembles the result of correlation clustering for this problem, as
\alg{$k$-means} tends to generate three clusters of nearly equal size. 

\begin{figure}[tb]
	\centering
	\subfigure{%
		\includegraphics[width=2in]{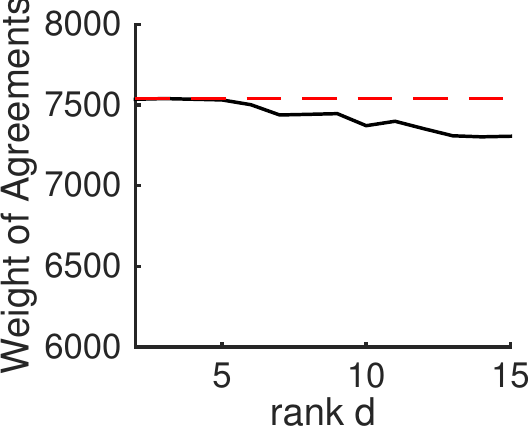}
		\label{fig:csobj1}} $\qquad$
	\subfigure{%
		\includegraphics[width=2in]{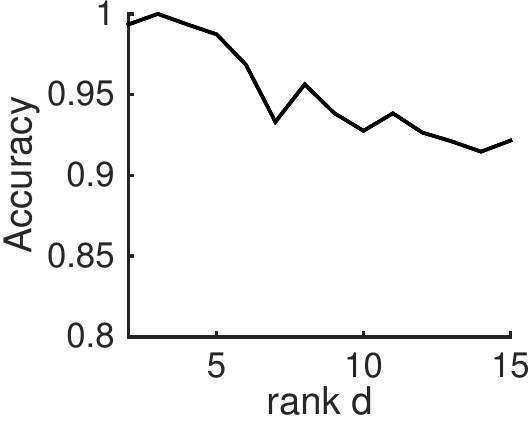}
		\label{fig:csobj2}}
	\caption{The left plot shows the objective scores achieved by
		\alg{ZonoCC} for CS conference search query data as we increase rank. The dotted line shows the score of the optimal clustering, which \alg{ZonoCC} remains very close to, regardless of
		rank. The right plot shows the accuracy of the rank-$d$ clustering, which is
		computed by considering all ${n \choose 2}$ pairs in the clustering, and
		finding the fraction of those pairs for which the decision to cluster together
		or cluster apart agrees with the optimal clustering of the dataset. In all
		cases, the accuracy of \alg{ZonoCC} is above 90\%.
	}
	\label{fig:2clus}
\end{figure}

We also run \alg{ZonoCC} for 50,000 iterations on rank-$d$ approximations of
the correlation clustering matrix, where~$d$ ranges from~$2$ to~$15$. As~$d$ increases, \alg{ZonoCC} tends to form more clusters, but is always able to identify two large groups of conferences that are highly correlated. In terms of the correlation clustering objective on the original (and not low-rank) matrix, \alg{ZonoCC} decreases in performance only because we must maintain a constant number of iterations for the sake of runtime, while the number of zonotope vertices increases exponentially in~$d$.
This behavior is illustrated in Figure~\ref{fig:2clus}:
even though these are sub-optimal, the clustering returned always has two large
clusters and small groups of outliers. 

\subsection{Stock Market Data}
\label{sec:stock}
The second study we consider is to cluster time series comprising
stock market closing prices on
different days of the year.
We obtain prices for~497 stocks from the S\&P 500
from Yahoo's Finance API over the~253 trading days in a
year.
We use the correlation between these time series to generate the input to our
correlation clustering experiment. In this case we are unable to run the ILP
to certify the clustering as optimal, and are unable to run the CGW algorithm
due to insufficient memory.
Thus, we just compare \alg{ZonoCC} against \alg{Pivot}: the resulting
clusterings are very close, but \alg{ZonoCC} finds the better clustering (see
the lower part of
Table~\ref{tab:cs} for the objectives and runtimes).

Similarly to our previous experiment, we discover that there are two large groups of closely correlated stocks and a third cluster with an outlier stock. The outlier is ``Public Storage'' (PSA), whose stock prices are largely uncorrelated with all other companies. For comparison, running \alg{$k$-means} with~3 clusters always splits up many of the closely correlated companies.  

\subsection{Clustering Networks via Embeddings}
\label{sec:embeddings}
We can use \alg{ZonoCC} to cluster any dataset where each entry is represented
by a vector in a low-dimensional vector space. This means that our algorithm
can be used to cluster unsigned network data as long as we have a way to embed
the nodes of the network in a low-dimensional space. Such embeddings have been an active area of research recently~\cite{grover2016node2vec,Perozzi-2014-deepwalk}. We demonstrate how to combine \alg{ZonoCC} with two different graph embedding techniques to produce large clusterings to analyze several networks from the Facebook 100 datasets~\cite{Traud-2012-facebook}.

The \emph{purpose} of these experiments is not to demonstrate that
\alg{ZonoCC} achieves a superior clustering result given the metadata.
(Indeed, there is no one algorithm that can achieve
this~\cite{Peel-2016-metadata}.) Rather we wish to compare
\alg{ZonoCC} and \alg{$k$-means} --
in terms of how their clusters reflect the metadata -- 
on low-quality embeddings from the eigenvectors of the
Laplacian and on a high-quality embedding from node2vec~\cite{grover2016node2vec}.

\textbf{Datasets}.
The datasets we use are subsets of the Facebook graph at certain US
universities on a certain day in 2005. These include an undirected graph and
anonymized metadata regarding each person's student-or-faculty status, gender, major, dorm/residence, and graduation year. We run our experiments on the following networks of different sizes: Reed, Caltech, Swarthmore, Simmons, and Johns Hopkins. We aim to cluster this data based on friendship links in the graph -- reflected in the embeddings -- and in the process see how the clusterings might be related to different attributes.  

\textbf{Embeddings}.
We consider two different ways to embed each node into a low-dimensional space based on the edge structure of the graph. The first is to take a subset of the eigenvectors of the normalized Laplacian of the network: $\mathcal{L} = \mathcal{I} - D^{-1/2}AD^{-1/2}$ where $D$ is the diagonal matrix of node degrees and $\mathcal{I}$ is the identity matrix. If we take the $d$ eigenvectors corresponding to the smallest nonzero eigenvalues of $\mathcal{L}$, this gives an embedding in $\mathbb{R}^d$.

The second embedding we consider comes from an algorithmic framework developed by Grover and Leskovec~\cite{grover2016node2vec} called \emph{node2vec} for mapping nodes in a network to a low-dimensional feature space for representational learning. The points of the embedding lie in $\mathbb{R}^d$ for a
user-specified~$d$. 

\textbf{Results}.
For each of the networks studied, we obtain two embeddings into
$\mathbb{R}^3$, one from the normalized Laplacian and the other using
\emph{node2vec}. For each embedding, we center the data by subtracting the
mean point. This gives us a set of $n$ vectors with both positive and negative
entries. We then run \alg{ZonoCC} on each embedding, and compare against
running \alg{$k$-means} for the same number of clusters as the output from \alg{ZonoCC}.

We analyze our clusterings by observing how the clusters relate to four of the meta-data attributes:
student-or-faculty status, major, dorm/residence, and graduation year.
{The metric we use is the proportion of pairs of people in the same cluster that share a given metadata attribute. Equivalently, we can think of this as the probability that two people selected uniformly at random from the same cluster share the attribute.}
We can also compute
this metric for the entire network to get a baseline score. The results for
this experiment are given in Table~\ref{tab:FB}, where the ``None'' method places
all nodes into a single cluster (which is the baseline probability). Note that
the only meaningful column across the networks is the \emph{Year} attribute. In addition, Caltech, which is a small school with a strong residential
population, shows a similar effect for the dorm attribute. Thus, we focus our attention on the Year attribute.

The table shows that for \emph{node2vec} embeddings, \alg{$k$-means} always gives a higher
proportion than \alg{ZonoCC} except for Caltech and Johns Hopkins, where they
are effectively the same. In contrast, for the embeddings from the Laplacian,
the \alg{ZonoCC} always shows stronger alignment with the year attribute. At
the very least, this is a demonstration that \alg{ZonoCC} and \alg{$k$-means} can alternate in performance on any given clustering task. However, we suspect that this is evidence that \alg{ZonoCC} is likely to be better in cases with \emph{weak features} (such as the Laplacian). 

\begin{table}[t!]
	\centering
	\fontsize{8}{9}\selectfont 
	\caption{Proportion of pairs of people in the same cluster that share the given attribute. All networks display a strong connection
		between the clusterings and the graduation year. \alg{ZonoCC} is better at
		detecting this trend on the low-quality Laplacian embedding, whereas \alg{$k$-means} performs better on the more sophisticated node2vec embedding.}
	\label{tab:FB}
	\begin{tabularx}{\linewidth}{@{}l@{\,}r@{\;\;}l XXXX@{}} 
		\toprule 
		Network& Emb. & Method & Stud.~or Fac. & Major & Dorm & Year\\ 
		\midrule 
		
		\textbf{Reed}& ---   & None  & 0.725 & 0.037 & 0.015 & 0.137 \\    \addlinespace
		$n =962$  &N2V & \alg{ZonoCC} &  0.698 & 0.039 & 0.018 & 0.278 \\ 
		&        & $k$-means& 0.756 & 0.039 & 0.020 & 0.325 \\    \addlinespace
		& Lap & \alg{ZonoCC} & 0.744 & 0.038 & 0.018 & 0.298 \\ 
		&     &  $k$-means & 0.745 & 0.038 & 0.018 & 0.290 \\ 
		\midrule
		\textbf{Caltech}&  --- & None  & 0.564 & 0.063 & 0.078 & 0.142 \\    \addlinespace
		$n = 769$ &N2V & \alg{ZonoCC} & 0.576 & 0.064 & 0.160 & 0.151 \\ 	
		&        & $k$-means& 0.566 & 0.065 & 0.127 & 0.145 \\    \addlinespace
		& Lap & \alg{ZonoCC} & 0.601 & 0.065 & 0.087 & 0.166 \\ 
		&     &  $k$-means & 0.578 & 0.064 & 0.080 & 0.146 \\ 
		\midrule
		\textbf{Swarthmore}& ---   & None  & 0.628 & 0.045 & 0.049 & 0.146 \\    \addlinespace
		$n = 1659$ &N2V & \alg{ZonoCC} & 0.620 & 0.048 & 0.055 & 0.262 \\ 
		&        & $k$-means& 0.627 & 0.049 & 0.051 & 0.265 \\    \addlinespace
		& Lap & \alg{ZonoCC} & 0.599 & 0.047 & 0.042 & 0.205 \\ 
		&     &  $k$-means & 0.599 & 0.046 & 0.042 & 0.197 \\ 
		\midrule
		\textbf{Simmons}& ---   & None  & 0.753 & 0.043 & 0.045 & 0.161 \\    \addlinespace
		$n = 1518$ &N2V & \alg{ZonoCC} & 0.716 & 0.043 & 0.064 & 0.378 \\ 
		&        & $k$-means& 0.717 & 0.043 & 0.065 & 0.379 \\    \addlinespace
		& Lap & \alg{ZonoCC} & 0.763 & 0.044 & 0.047 & 0.167 \\ 
		&     &  $k$-means & 0.761 & 0.044 & 0.045 & 0.162 \\ 
		\midrule
		\textbf{Johns Hop.}& ---   & None  & 0.618 & 0.036 & 0.020 & 0.134 \\    \addlinespace
		$n = 5180$ &N2V & \alg{ZonoCC} & 0.612 & 0.041 & 0.025 & 0.213 \\ 
		&        & $k$-means& 0.592 & 0.041 & 0.024 & 0.203 \\    \addlinespace
		& Lap & \alg{ZonoCC} & 0.632 & 0.046 & 0.026 & 0.229 \\ 
		&     &  $k$-means & 0.608 & 0.042 & 0.023 & 0.187 \\ 
		\midrule
	\end{tabularx}
\end{table}	

\section{Related Work}
Our work builds on several years of research in correlation clustering. The
problem was originally introduced for complete and unweighted graphs by Bansal, Blum, and Chawla~\cite{bansal2004correlation}, who proved NP-hardness and gave a PTAS
for maximizing agreements and a constant factor approximation for minimizing
disagreements. Charikar, Guruswami, and Wirth~\cite{charikar2003clustering} later extended
these results by improving the constant factor approximation for minimizing
disagreements, and gave a 0.7664-approximation for maximizing agreements in general weighted graphs based on a semidefinite programming relaxation. The approximation for minimizing disagreements in unweighted graphs was improved to~2.5 by Ailon, Charikar, and Newman~\cite{ailon2008aggregating}, who at the same time developed the simplified \alg{Pivot} algorithm. 
Recently Asteris et al.~\cite{asteris2016bipartite} gave a PTAS for maximizing agreements on unweighted bipartite graphs by obtaining a low-rank approximation of the graph's biadjacency matrix. Our work extends these results by showing how low-rank structure can also be exploited for general weighted correlation clustering.

Beyond the correlation clustering literature, our work shares similarities with other results on NP-hard problems that become solvable in polynomial time for low-rank positive semidefinite input. Ferrez, Fukuda, and Liebling gave a polynomial time solution for maximizing a  quadratic program in $\{0,1\}$ variables on low-rank positive semidefinite matrices~\cite{ferrez2005solving}, and Markopoulos, Karystinos, and Pados proved an analogous result for the $\pm 1$ binary case~\cite{markopoulos2014optimal}. While these results seek to optimally partition a set of vectors into two clusters, our work can be seen as a generalization to arbitrarily many clusters.

Our approach in solving low-rank correlation clustering shares many similarities as well with the \emph{spannogram} framework for exactly solving combinatorially constrained quadratic optimization problems on low-rank input~\cite{karystinos2010efficient,asteris2014sparse,papailiopoulos2013sparse}. In particular, for the NP-hard densest subgraph problem, Papailiopoulos et al.~used this framework to prove that a low-rank bilinear relaxation of the densest subgraph problem is solvable in polynomial time for low-rank input~\cite{papailiopoulos2014finding}. 

\section{Conclusions}
Our results introduce a new approach to solving general weighted correlation clustering problems by considering the rank and structure of the underlying matrix associated with the problem. This opens a number of new directions in correlation clustering-based approaches. The algorithm  we present offers a fast and accurate method for solving correlation clustering problems where the input can be represented or at least well approximated by a low-rank positive semidefinite matrix.  We demonstrate a number of applications including clustering time series data from search queries relating to top-tier computer science conferences and stock closing prices. We also demonstrate how this method can be used with embeddings of network data into low-dimensional spaces.

In future work we wish to prove more rigorous theoretical approximation results for our methods. Specifically, we would like to give an approximation bound for $k$ iterations of \alg{ZonoCC}, and also give rigorous bounds on the correlation clustering objective when taking a low-rank approximation.

\section{Acknowledgments}
Nate Veldt is funded by NSF award IIS-1546488, and would like to thank the NSF and the Australian Academy of Science for jointly funding his work as a part of the East Asia and Pacific Summer Institute in June-July 2016 (NSF award OISE-1613938), during which time much of the above work was performed.
Anthony Wirth thanks the Australian Research Council (Future Fellowship FT120100307)
and the Melbourne School
of Engineering for funding visits by him to Purdue and David Gleich to
Melbourne. David Gleich is partially supported by NSF CAREER award CCF-1149756, NSF award IIS-1546488, NSF STC award CCF-093937, the DARPA SIMPLEX program, and the Sloan Foundation.

\newpage
\bibliographystyle{abbrv}
\bibliography{fullbib}

\end{document}